# A review of ontologies for smart and ongoing commissioning

Sara Gilani, Caroline Quinn, J.J. McArthur*

Faculty of Engineering and Architectural Science, Ryerson University, Toronto, Canada

* Corresponding Author: 325 Church Street, Toronto, ON M5B 2K3, Canada; Tel: 1 416 979 5000 x554082; Fax: 1 416 979 5353; jjmcarthur@ryerson.ca.

**Abstract**

Smart and ongoing commissioning (SOCx) of buildings can result in a significant reduction in the gap between design and operational performance. Ontologies play an important role in SOCx as they facilitate data readability and reasoning by machines. A better understanding of ontologies is required in order to develop and incorporate them in SOCx. This paper critically reviews the state-of-the-art research on building data ontologies since 2014 within the SOCx domain through sorting them based on building data types, general approaches, and applications. The data types of two main domains of building information modeling and building management system have been considered in the majority of existing ontologies. Three main applications are evident from a critical analysis of existing ontologies: (1) key performance indicator calculation, (2) building performance improvement, and (3) fault detection and diagnosis. The key gaps found in the literature review are a holistic ontology for SOCx and insight on how such approaches should be evaluated. Based on these findings, this study provides recommendations for future necessary research including: identification of SOCx-related data types, assessment of ontology performance, and creation of open-source approaches.

**Nomenclature**

| Ontology Abbreviation | Full name |
|---|---|
| BACS | Building Automation and Control Systems Ontology |
| BIO | gbBuilding Information Ontology |
| BOT | Building Topology Ontology |
| bSDD | buildingSMART Data Dictionary Ontology |
| CIM | Common Information Model Ontology |
| DUL | DOLCE Ultralite Ontology |
| EEPSA | Energy Efficiency Prediction Semantic Assistant Ontology |
| ESIM | Energy System Information Model Ontology |
| FIEMSER | Friendly Intelligent Energy Management Systems in Residential Buildings Ontology |
| FSM | Finite State Machine Ontology |
| FUSE-IT | Facility Using Smart secured Energy and Information Technology |
| HTO | Haystack Tagging Ontology |
| IDS | Intrusion Detection System Ontology |
| MAMO | Mathematical Modeling Ontology |
| OM | Units of Measure (OM) Ontology |
| OntoMODEL | Ontological Mathematical Modeling Knowledge Management Ontology |
| OSPH | Object States and Performance History |
| OWL-S | OWL for Web Services |
| PO | User Behavior and Building Process Information Ontology |
| QUDT | Quantities, Units, Dimensions and Types Ontology |
| SAREF | Smart Appliances REFerence Ontology |
| SBMS | Semantic BMS Ontology |
| SEAS | Smart Energy Aware Systems Ontology |
| SimModel | Simulation Domain Model Ontology |
| SOSA | Sensor-Observation-Sampling-Actuator Ontology |
| SSN | Semantic Sensor Network Ontology |
| UCO | Unified Cybersecurity Ontology |
| UEO | Urban Energy Ontology |
| WGS84 | WGS84 Geo Positioning Ontology |
| WO | Weather Ontology |

# 1 Introduction

"Commissioning" of a building is a process to test that a building operates as expected [1] and resolving any performance gaps to improve users' comfort in buildings [2, 3, 4]. Such discrepancies between design and operational-phase energy use of buildings is widely recognized [5]. When the commissioning process is used throughout a building's life cycle, "ongoing commissioning" will perpetually verify whether a building meets its intended performance. Through ongoing commissioning, new issues such as changes in buildings' use and energy systems can be identified in operation, improving risk management [6], and reducing building energy use [7, 6] and cost [3]. With the emerging sensor technologies, availability of real-time data has provided opportunities to incorporate "smart and ongoing commissioning" (SOCx). Similar to ongoing commissioning in providing the potential for energy and cost savings, smart commissioning using wireless sensor network has shown promising potential for energy and cost reduction [8, 9, 10, 11] due to changes in the occupancy, operating hours, and internal loads [3].

Implementation of SOCx necessitates real-time data streaming and analytics. However, main challenges in this regard are the heterogeneity and volume of data. Combining data from different information resources necessitates extensive human labor given the heterogeneous nature of the vast amounts of data collected using various systems incorporated in the Architecture, Engineering, Construction, and Operation (AECO) industry [12]. To deal with these challenges, semantic web technologies have shown great potential. These technologies provide data webs where the data are open, reusable, extensible, and machine readable [13]. Sensor-based data can be processed automatically using semantic web technologies where they can also be published in the open data cloud to be used in any application [14]. The usage of semantic web technologies in buildings has increased significantly in the past decade, as SOCx requires data management across different building-related domains, such as building information modeling (BIM) and building management system (BMS). Semantic web technologies have also enabled cross-domain interoperability in the Internet of Things (IoT) [14, 15].

Another key element to support SOCx is the need for an *ontology* to support data readability and machine reasoning [16]. There are various definitions for ontologies [17]. Gruber [18] described an ontology as a means to explicitly define concepts, classes, and objects and the corresponding relationships in an area of interest. Weise et al. [19] defined an ontology as a formal model to represent the knowledge of a specific domain, whereas Guarino [20] classified different types of ontologies based on their generality level. Guarino [20] categorized ontologies as top-level, domain, task, and application. Top-level ontologies define general concepts irrespective of domains of interest. Domain ontologies define specific concepts and vocabularies related to a specific domain, while task ontologies describe specific vocabularies related to a specific task or activity. Application ontologies define concepts related to both domain and task ontologies. Guarino's [20] hierarchical approach considered ontologies as modules that can be general and expandable [21].

Using ontologies in the building domain has attracted much attention in recent years. Several previous studies on building data ontologies have either reviewed or analyzed existing ontologies comparatively. For instance, Bhattacharya et al. [22] provided a comparative analysis of three ontologies related to building sensor data and building information including Haystack [23], SAREF [24], and an ontology defined by the Industry Foundation Classes (IFC) schema, and tested the performance of these three ontologies in case studies. Borgo et al. [25] critically studied the IFC schema as a standard data model commonly used in the AECO industry to see how it can be converted into an Ontology Web Language (OWL) format to make it compatible with other semantic models. Likewise, Gerrish et al. [26] explored the capabilities of the IFC schema to integrate a BIM model with building energy models and BMS data for building performance management.

Bajaj et al. [27] reviewed existing ontologies developed in the IoT domain since 2012. Their objective was to identify the useful concepts for developing a unified IoT ontology and to structure the existing IoT-related literature to help developers create a unified standard ontology. Butzin et al. [16] also reviewed existing BMS and sensor-related ontologies to identify the information models used to define BMS entities, devices, and contextual information. They explored the information domains that were included, the vocabularies used in BMS devices and systems, and the structure of BMS ontologies. A recent study by Fierro and Culler [28] specifically focused on the Brick ontology [29] to develop a query method for fast Brick querying for the applications sensitive to time (such as model predictive controls and demand response).

While the aforementioned papers studied a single or cross-domain ontology relevant to buildings, there remains a gap in the literature regarding a comprehensive review of the state-of-the-art ontologies within the SOCx domain. To address this gap, this paper critically reviews and identifies the characteristics of existing ontologies within the SOCx domain.

The focus of this research was on the ontologies at the building scale rather than city scale. The recent papers on building data ontologies rather than architecture and framework (for which the reader is recommended to refer to [30, 31, 32, 33, 34, 35, 36, 37, 38]) were in the scope of this study.

The main research questions of this literature review paper were:

1. What building data types have been included in existing ontologies?
2. What are the general approaches of existing ontologies?
3. What are the applications of existing ontologies?
4. What are the general trends and gaps of existing ontologies?

These questions form the structure of the main body of the paper, which concludes with an overall discussion of general trends and insight on future research needs in this field.

## 2 Literature search process

To conduct a comprehensive literature review on the data ontology for SOCx, the Preferred Reporting Items for Systematic Reviews and Meta-Analyses (PRISMA) process [39] was used. This process requires a systematic narrowing of relevant publications given a set of input search terms and databases through four main phases: identification, screening, eligibility, and included (Figure 1).

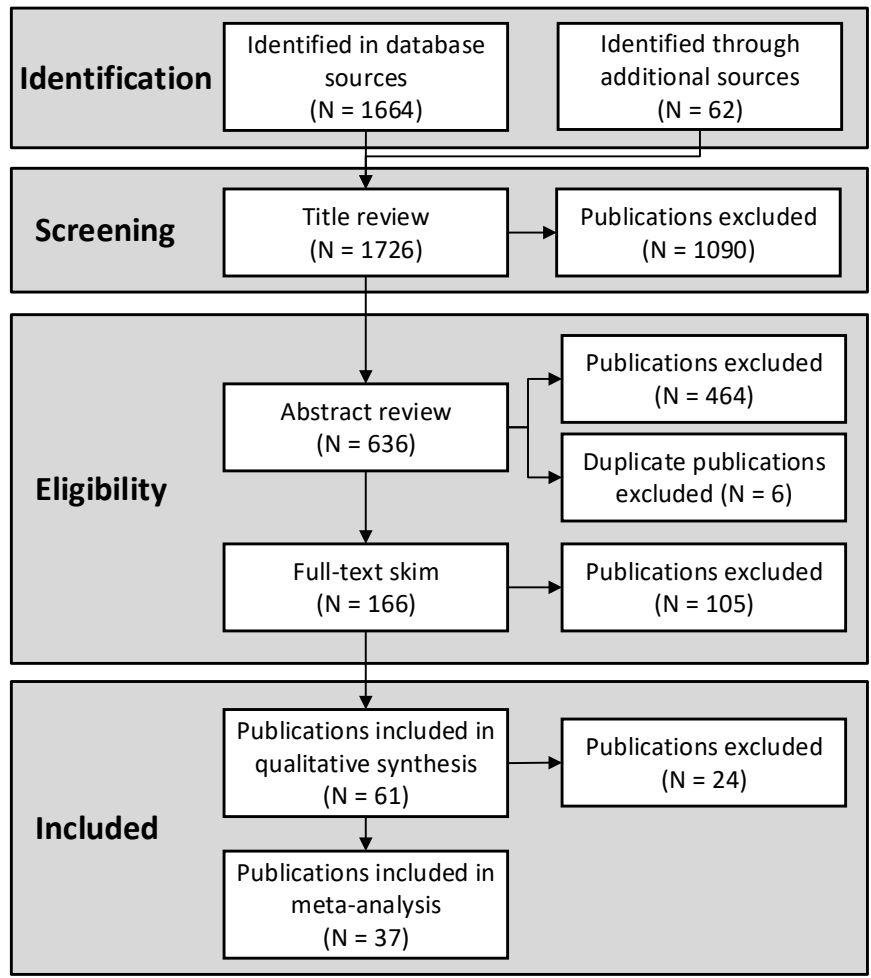

*Figure 1. PRISMA process flow.*

The PRISMA process allows for an interdisciplinary literature review, as well as the review of a topic that is not the focus of publications but is discussed within them. This exhaustive process is well-suited for this literature review because the topic is vast and studied across multiple disciplines.

To further ensure that the topic was sufficiently covered, primary search terms and synonyms were used. The primary search terms identified were: ontology, building, and commissioning. The following logic was input into the search bar of various databases to capture ontology-related terms (ontology, framework, information model), building domain (building), and commissioning-related terms (commission, energy management, facility management): {ontology AND building AND {commission OR energy management OR facility management}} OR {framework AND building AND {commission OR energy management OR facility management}}, OR {information model AND building AND {commission OR energy management OR facility management}}. The search terms were entered into the six databases: ACM [40], IEEE [41], Inderscience [42], SciX [43], Scopus [44], and Taylor and Francis [45]. Several other publications found in the initial topic search were included in a category titled "Additional" in the identification phase of the PRISMA process.

Table 1 demonstrates the four phases of the PRISMA process for the current literature review study. In the identification phase, any publication meeting the criteria of: (1) containing search terms in the title, abstract, or keywords, (2) available in English, (3) written in the last five years (since 2014), and (4) peer reviewed, was added to the list for further consideration in the next phase of the PRISMA process. In the screening phase, titles of publications were manually screened. If the title of a publication indicated the topic of data management but did not relate to the AECO industry, the publication was flagged as irrelevant. In the eligibility phase, abstracts of the publications were reviewed in a first pass and the full text was skimmed in a second pass. In each pass, if it was discovered that the abstract or full text did not meet the criteria of the screening phase, it was removed from the working list. Furthermore, in the eligibility phase, publications overly focused on computer science or algorithmic data calculation, discussing residential buildings, or pertaining to smart grid or smart city technology were considered irrelevant. To determine which publications should be included, full publications were read and those without detailed descriptions of building ontology, or those overly focused on data architecture or framework development were excluded from the meta-analysis, but included in the qualitative analysis of the literature review. Qualitative analysis involved defining data ontology within the context of SOCx and helped inform the discussion section of this paper, whereas meta-analysis answered research questions.

Table 1 shows the number of papers selected from each database at each step in the PRISMA process. The IEEE and Scopus databases gave the most publications within the scope of the review, and the Inderscience database the least. In a few instances, publications appeared on both IEEE and Scopus were represented in IEEE, as overall this database offered a greater number of relevant papers to the research topic.

Table 1. Number of publications selected by database at each phase of the PRISMA process.

| Source | Identification | Screening | Eligibility | Included |
|---|---|---|---|---|
| ACM | 528 | 140 | 28 | 2 |
| IEEE | 227 | 115 | 38 | 15 |
| Inderscience | 16 | 3 | 0 | 0 |
| SciX | 20 | 14 | 8 | 2 |
| Scopus | 840 | 285 | 59 | 9 |
| Taylor and Francis | 33 | 17 | 5 | 1 |
| Additional | 62 | 62 | 28 | 8 |
| **Total** | **1726** | **636** | **166** | **37** |

Based on the PRISMA process, in total, 37 papers were selected and reviewed critically for the meta-analysis of the current study.

## 3  Building data types

Building data ontologies employ data from a variety of sources in a building. These sources offer varied measures and data points that can be grouped by themes into data types. This allows for the comparison of ontologies by data type. Of the selected 37 papers, 31 were considered for the ontology data type analysis, as the remaining six papers were reviews or comparison of existing ontologies and provided inadequate information for such a comparison.

First, the study categorized building data as either static or dynamic and identified which type(s) were used in each ontology. Static data types describe physical building information, whereas dynamic data types involve time-series data [46]. Ontologies with both static and dynamic data types were the most common, representing 90% of analyzed ontologies, while the remaining 10% used only static data types. No ontologies considered dynamic data only. Those ontologies using only static data types are still relevant to SOCx because they either discussed linking simulation-based data to physical building information [47, 12] or discussed how BMS elements can be mapped to physical building information [37].

Further to the initial static and dynamic data type categorization, on the basis of existing ontologies [48, 49, 29] with more comprehensively defined data types in the literature, the current study categorized data types required for a SOCx ontology as: (1) external conditions, (2) indoor conditions, (3) building systems and components, (4) energy use, (5) maintenance, (6) occupant-related data, (7) physical building information, (8) performance-based data, and (9) simulation-based data, as presented in Table 2.

*Table 2. Summary of data types used in state-of-the-art ontologies.*

| Data type | Reference |
|---|---|
| External conditions | [50], [14], [21], [51], [52], [53], [54], [49], [55], [35], [56], [57], [58], [59], [60] |
| Indoor conditions | [61], [62], [63], [50], [14], [48], [24], [21], [47], [51], [64], [52], [15], [65], [54], [49], [37], [55], [35], [56], [29], [57], [58], [15], [66], [67], [68], [59], [69], [70], [60] |
| Building systems/components | [61], [62], [63], [50], [14], [24], [21], [47], [51], [64], [15], [65], [54], [49], [55], [35], [56], [29], [57], [58], [66], [68], [59], [69], [70] |
| Energy use | [62], [50], [14], [48], [24], [21], [51], [52], [53], [65], [54], [49], [55], [35], [56], [29], [58], [66], [59], [69], [60] |
| Maintenance | [53], [65], [54], [68], [69], [70] |
| Occupant-related data | [62], [14], [48], [24], [21], [51], [52], [53], [65], [49], [55], [57], [68], [59], [70], [60] |
| Physical building information | [61], [62], [14], [48], [21], [47], [51], [64], [52], [15], [65], [54], [49], [37], [55], [35], [56], [29], [57], [58], [15], [66], [67], [68], [59], [69], [70] |
| Performance-based data | [48], [21], [47], [51], [64], [53], [58], [59], [69], [70], [60] |
| Simulation-based data | [62], [63], [48], [47], [51], [55], [68], [59] |

In addition to building data, building performance evaluation requires an understanding of climatic conditions, which are either measured or obtained from weather forecasts. Knowledge of indoor environmental conditions measured by BMS-integrated sensors are also important since associated data points are used as input to a BMS for adjusting actuators. The status of building systems and components, which may or may not be integrated with a BMS, are required to both control their functionalities and track occupants' interactions with buildings. Energy use monitoring of building systems and spaces at different temporal scales are required to verify building energy performance. The maintenance history of malfunctioning building systems and components and corresponding repairing actions are other data types desired for a SOCx ontology to support facility management activities. In addition to building-related data as critical elements in building operation, information about occupants is also imperative in order to provide a comfortable space for a building's occupants while reducing energy use. Furthermore, the use of physical building information provides static spatial data including the building model, sensor locations, and HVAC component locations as well as building envelope properties. While the aforementioned data types are monitored in a building, there remains two more data categories recognized in the literature [48, 58] typically considered as virtual data types: performance and simulation-based data. Performance-based data are the data scored using key performance indicators (KPIs) as measures to evaluate building performance (energy and occupants' comfort). Simulation-based data are outputs resulted from simulation-aided process (e.g. to assess retrofitting solutions) or other calculation methods (e.g. equations for predicted mean vote).

Table 2 indicates that among the building data types in building ontologies, the common data types included were indoor conditions and physical building information. With the emerging BMSs and BMS-integrated sensors, the use of BMS as a centralized building management and operation system has become prevalent to record environmental conditions. Likewise, physical building information data are critical data types for a building ontology using the well-established schema such as IFC in the AECO industry to define the spatial relationship between data points. On the other hand, maintenance data was the least considered data type in a building ontology, although it is highly relevant to SOCx. This is partially due to the lack of agreed methods to extract semantic knowledge from unstructured maintenance textual information such as work orders [71], suggesting that there is a research gap in this area for SOCx ontology development.

## 4     General ontological approaches

Across the literature reviewed, a number of approaches for ontology-related publications were noted. These approaches can be categorized as: (1) original ontology creation, (2) synthesis of existing ontologies, and (3) review and comparative analysis of existing ontologies.

Originally developed ontologies are generally case specific. For instance, the SPORTE2 ontology [62] was developed for a specific case study (a swimming pool hall) to actuate the supply air temperature and flow rate to the space and reduce its energy use, for which main concepts were scenario definitions obtained through pilot studies rather than general concepts. Likewise, Mallak et al. [68] developed an ontology to detect HVAC systems' faults in a specific office building. Kaed et al. [53] developed an entirely new ontology (extended in a later work [15]) to link BMS data types with other domains with the constraints of proprietary systems deployed in a specific building. Hence, these case study-based ontologies are unlikely to be applicable to other buildings, especially with proprietary systems.

At present, originally developed ontologies have been mostly created to manage a single-domain data types. Shamszaman et al.'s [50] ontology and Charpenay's [14] HTO ontology mainly focused on BMS data. Jeong and Son [47] developed an original ontology to convert BIM data to an object-oriented physical model which could be used by object-oriented modeling languages (e.g. Modelica [72]) for building performance simulation and parametric analysis. Similarly, the BOT ontology [37] was focused on physical building information, however unlike Jeong and Son's [47] ontology where BIM environment was used to define a building physical model, the BOT ontology defined the core topology of a building model in which a building was considered as the physical and conceptual objects and the associated relationships.

On the multiple-domain data types front, the cloud-hosted usage of ontologies has yet to be tested as the current originally developed ontologies are in the form of a framework. For instance, Mahdavi and Taheri [49] provided illustrative examples of representing monitored data (window position, indoor temperature, outdoor relative humidity, valve position, operational state of office equipment, and electricity energy consumption) for the six categories disaggregated to occupants, indoor conditions, outdoor conditions, control systems and devices, equipment, and energy flows, regardless a cloud-based platform over the web. Schachinger and Kastner [55] also developed an original ontology for a knowledge-based framework using machine-readable semantics to host the required information for building optimization, however they discussed its functionality only from the data retrieval perspective rather than online optimization in a case study.

Implementation of wholly new ontologies in contexts that are different from their original contexts also requires high level of time, effort, and evaluation. For instance, while the Brick ontology [29] has enriched the location concept of the IFC schema and the tagging concept of Haystack [23] to structure building data for building operation through modeling hierarchies and relationships, mapping existing buildings' raw metadata and their relations to Brick has not yet been fully automated, thus not scalable at present. However, artificial intelligence (AI) can be used to automate this process in future work. For example, recent studies have demonstrated promising results of using various machine learning techniques for automated labeling of BMS data points [73, 74] and inference of relationships between sensors and equipment [75, 76].

While original ontology creation imposes the aforementioned limitations, synthesis of existing ontologies enhances the quality of ontologies and reduces time and cost in developing and maintaining ontologies. Multiple building data models, covering both static and dynamic data types, can be combined in a single ontology. For instance, Asfand-e-yar et al. [61] synthesized existing ontologies from two domains of BIM and BMS to manage location and device information and read BMS data. InterfaceOnto [51] is also an ontology to provide an interface between BIM and BMS data. Similarly, the BACS ontology for building controls [56] and the ESIM ontology for building performance improvement [69] combined BIM and BMS data. The OPTIMUS ontology [59] also exemplified the possibility of using a large number of data types, reusing existing ontologies.

Using existing ontologies also enables ontology developers to extend existing ontologies. For instance, the SBMS ontology extended the SSN ontology for BMS data to combine BMS data with location and device information independent of the automation protocol using the SBIM ontology [66]. Furthermore, using existing ontologies facilitates designing modular ontologies. For example, the SAREF ontology [24] for semantic interoperability between various smart appliances was developed based on the modularity principle to facilitate different combinations of various parts of it, reusing existing concepts and relationships within the domain.

Table 3 summarizes general approaches of the state-of-the-art publications on ontologies. As this table indicates, the majority of the state-of-the-art ontologies, especially in recent years, have been developed based on reusing existing ontologies for various building data types. This direction in developing ontologies indicates that there are a large number of existing ontologies that can be tailored to meet SOCx applications rather than creating ontologies from scratch.

*Table 3. General approach of state-of-the-art publications on ontologies.*

| Approach | Reference |
| --- | --- |
| Original ontology creation | [62], [50], [14], [47], [53], [49], [37], [55], [29], [15], [68] |
| Synthesis of existing ontologies | [61], [63], [48], [24], [21], [51], [64], [52], [65], [54], [35], [56], [57], [58], [66], [67], [59], [69], [70], [60] |
| Review and comparative analysis of existing ontologies | [25], [22], [26], [27], [16], [28] |

The common building data types for which existing ontologies were reused included: indoor conditions and building systems/components, physical building information, simulation-based data, and other building data such as building location, as presented in Table 4. Since the extraction of some data types (e.g. performance-based data and control actuators) from BMS-integrated sensors requires using calculation methods, existing ontologies (under so-called mathematical-based data in Table 4) were also reused in several cases. Moreover, there were general concepts (such as time and unit) that were commonly used to identify attributes of interest, regardless of building data types. The relevant reused ontologies in these cases were classified under top-level ontologies in Table 4 based on Guarino's [20] definition.

*Table 4. Common data types for existing ontologies named and reused in state-of-the-art ontologies.*

| Data type | | Existing ontology | Reference |
|---|---|---|---|
| Indoor conditions and building systems/components | | FSM | [56] |
| | | SAREF | [65], [70] |
| | | SEAS | [65], [35], [70] |
| | | SOSA | [56], [57] |
| | | SSN | [48], [52], [65], [56], [57], [58], [59], [70], [60] |
| | | VDI Guideline | [67] |
| Physical building information | | BIO | [60] |
| | | BOT | [56], [57] |
| | | bSDD | [64] |
| | | FIEMSER | [24] |
| | | IFC schema | [61], [64], [65], [54], [69], [70] |
| | | ifcOWL | [48], [51], [58] |
| | | OWL | [57] |
| | | UEO | [59] |
| Simulation-based data | | SimModel | [63], [48] |
| Other building-related data | Mathematical-based data | express | [56] |
| | | ifcmr | [56] |
| | | list | [56] |
| | | MAMO | [60] |
| | | OntoMODEL | [60] |
| | | OpenMath | [58] |
| | | OWL-DL | [21] |
| | | statistics | [56] |
| | Building location | WGS84 | [24], [21], [60] |
| Top-level concepts | | DUL | [52], [60] |
| | | OM | [35], [60] |
| | | OWL-time | [24], [21], [52], [60] |
| | | QUDT | [57] |

Within the literature, there were several ontologies reused frequently by ontology developers. The two main domains for which existing ontologies were frequently reused were BMS (indoor conditions and building systems/components) and physical building information. For instance, in most cases, existing ontologies defined by the SSN ontology and IFC schema were reused for the BMS data and physical building information, respectively. The use of these two dominant domains is due to the fact that real-time data streaming necessitates BMS data and the physical building information is required for context.

This review identified several very specific ontologies that were used only by a few researchers, though there were thematic clusters among them. For instance, only two of the reviewed ontologies [65, 60] reused information and communications technology (ICT) (e.g. IDS [77], UCO [78], OWL-S [79]), weather (e.g. WO [80]), micro-grid systems (e.g. CIM [81, 82]), and occupant-related ontologies (e.g. PO [83], DNAs [84, 85]). This trend of incorporating rarely reused ontologies indicates that while there are building data types (e.g. cyber-security and occupant-related data) that are highly beneficial for SOCx, there is still low penetration of well-established ontologies for these data types into practice. This is partially due to the scarcity of existing open-source and extensible ontologies for these data types, which necessitates future research.

## 5 Application of ontologies

An effective SOCx ontology requires a coherent data structure to model all building data types. The data types of the two domains of physical building information and BMS have been included in the majority of reviewed ontologies (as discussed in Section 3). A BMS monitors environmental conditions dynamically; and several previous ontologies specifically focused on these data types such as defining general BMS concepts (Haystack model [23], HTO [14], and SAREF ontology [24]), creating a semantic BMS data structure [50], or dealing with multi-technology systems [67]. On the static data front, a building information model provides this data; in this respect, some ontologies attempted to define the core topology of a building model [37] or provide building simulation data models [63, 47, 64] to deal with the BIM data.

Some ontologies also tried to unify BMS and BIM data in a single ontology. For example, Asfand-e-yar et al. [61] synthesized the IFC schema to manage BIM data and the communication protocol BACnet for BMS data to retrieve information of interest effectively for building operators and facility managers at large scale (a university campus). The Brick ontology [29] also combined the location concept of the IFC schema for BIM data with the tagging concept of Haystack [23] for BMS data. Another example is the SBMS and SBIM ontology [66] where BMS data was enriched through linking BMS with BIM data, and semantic information were provided for building operators (using a SPARQL query processing).

Beyond combining BMS and BIM data, several ontologies linked BMS data types with other building-related domains such as power monitoring system and electrical and gas utility billing information [53, 15], energy supply and efficiency, building automation, ICT, and security and safety in the FUSE-IT ontology [65], and occupant-related data [49].

The key trends of reviewed ontologies show that inclusion of cross-domain data types in a SOCx ontology is important. For instance, BIM data are essential for location and device information and BMS data for system and environmental information [61, 66]. ICT, safety and security [65], and occupants-related data [49] have also emerged as necessary data in developing ontologies within the SOCx domain due to the prevalence of ICT technologies, vulnerability of critical sites to security threats, and significant role of occupants on buildings' performances, respectively. To combine these various data types, the Resource Description Framework (RDF) schema was incorporated in several ontologies [63, 14, 29, 66], as this schema based on the triple concept (subject–predicate–object) effectively supports describing an ontology's classes and the relationships between them for knowledge representation. For retrieving data from the RDF schema, SPARQL was used as an efficient query language to reflect relationships in most case studies of the reviewed ontologies [61, 29, 66].

Beyond defining general data types and a data structure for a SOCx ontology, knowledge specific to this area must also be represented in a machine and human readable format. A SOCx ontology developer should therefore be cognizant of relevant applications. In recognition of this, the main applications of the 31 reviewed ontologies were explored. Analysis of these ontologies revealed three main applications: (1) KPI calculation, (2) building performance improvement, and (3) fault detection and diagnosis (FDD), as summarized in Table 5.

*Table 5. Main applications of state-of-the-art ontologies.*

| Application | Reference |
| --- | --- |
| KPI calculation | [48], [58], [60] |
| Building performance improvement | [62], [21], [51], [52], [54], [55], [56], [57], [69], [59] |
| Fault detection and diagnosis | [35], [68], [70] |

The state-of-the-art ontologies with the three aforementioned applications are discussed in this section, along with their advantages and limitations as summarized in Table 6**Error! Reference source not found.**.

*Table 6. Comparison of state-of-the-art ontologies with KPI calculation, building performance improvement, and FDD applications.*

| Reference | Advantage | Limitation |
| --- | --- | --- |

| Ref | Including multiple data types | Reusing existing ontologies | Using a modular design pattern | Tested in case study at building level | Using a context-aware design pattern | Streaming real-time data | Adjusting actuators | Developing wholly new ontology | Providing a conceptual design | Limited to rule-based control model | Tested in case study at zone level | Not tested in case study in operation | Redefining classes (not using predefined classes) |
|---|---|---|---|---|---|---|---|---|---|---|---|---|---|
| [48] | ✓ | ✓ | ✓ | | | | | | | | ✓ | | |
| [58] | ✓ | ✓ | | | | ✓ | | | | | ✓ | | |
| [60] | ✓ | ✓ | | ✓ | | | | | | | | | |
| [62] | ✓ | | | | | ✓ | ✓ | ✓ | | | | ✓ | |
| [21] | ✓ | ✓ | ✓ | | ✓ | | | | | | ✓ | | |
| [51] | ✓ | ✓ | ✓ | | ✓ | | | | ✓ | | | ✓ | |
| [52] | ✓ | ✓ | | ✓ | ✓ | | | | | | | | |
| [54] | ✓ | ✓ | | | | ✓ | | | | | ✓ | | |
| [55] | ✓ | | | | | ✓ | ✓ | ✓ | | | ✓ | | |
| [56] | ✓ | ✓ | | | | ✓ | ✓ | | | ✓ | ✓ | | ✓ |
| [57] | ✓ | ✓ | | | | | | | | | ✓ | | |
| [69] | ✓ | ✓ | | | | | | | | | | ✓ | |
| [59] | ✓ | ✓ | | | | | | | | | ✓ | | |
| [35] | ✓ | ✓ | ✓ | | ✓ | ✓ | | | | ✓ | ✓ | | ✓ |
| [68] | ✓ | | | | | ✓ | | ✓ | | ✓ | ✓ | | |
| [70] | ✓ | ✓ | | | | ✓ | ✓ | | | | ✓ | | |

## 5.1. KPI calculation

While managing building data is a fundamental stage in developing a SOCx ontology, consensus on what makes a building energy efficient and comfortable is also required. Thus, KPI calculations have emerged to assess the performance of buildings across various considerations in SOCx, and this is reflected in several reviewed ontologies [48, 58, 60].

Corry et al. [48] developed an ontology for building performance assessment throughout the life-cycle of buildings to reduce the gap between the real and predicted building performance through defining performance objectives and a set of related evaluation metrics. To achieve this goal, the simulated and measured KPIs of building performance were compared and tested at the zone level (a computer suite in an academic building in Ireland) to evaluate the thermal comfort conditions and HVAC performance using the ontology that was driven by sharing semantic data over the network. For instance, the gap between the simulated and measured predicted mean vote was reduced through the adjustment of the measured room temperature and assumed activity level of occupants on the basis of the data queried using Corry et al.'s [48] KPI ontology. This ontology used a modular data structure, as opposed to monolithic data structure, to enhance its extensibility and maintainability. However, while this ontology used semantic web technologies to describe sensors and the relationships between them, it did not support real-time data streaming.

In an attempt to automate this process, Hu et al. [58] created a KPI ontology by developing two main algorithms: one for preparing data streaming from multiple sources, and one for computing building performance. The implementation of the proposed ontology through integrating OpenMath and Linked Data using the RDF schema and SPARQL query language in a case study (a pool hall of an academic building in Ireland) provided insights into changing an HVAC control strategy without affecting thermal conditions negatively. Unlike Corry et al.'s [48] ontology, Hu et al.'s [58] ontology provided time-series data analysis. As the time-series data were decoupled from the building performance calculation, they could be reused for other purposes such as building operation. This arrangement provided flexibility in using time-series data in future throughout the building life cycle. Furthermore, this ontology linked contextual building information with KPI calculation, simplifying meaningful information extraction from various building domains.

At a multi-scale level, Li et al. [60] developed a KPI ontology to facilitate building and district performance tracking. The implementation of this ontology eased the assessment of energy balance and time correlation at both building and district levels. Among the reviewed KPI ontologies, Li et al.'s [60] ontology was tested at a relatively large scale. The feasibility of this ontology was demonstrated using the dataset from the Solar Decathlon Europe 2012 competition in Spain, where 19 solar houses were connected to a microgrid. This ontology combined a considerable number of existing ontologies to cover data required for performance assessment from the building to district scale.

The reviewed KPI ontologies benefited from reusing existing ontologies, promoting their extensibility and scalability. For example, the ifcOWL ontology was used to convert the IFC building information model to RDF graphs, the SimModel ontology was used to create an interoperable XML-based building data model for simulating the building (e.g. using EnergyPlus, OpenStudio) [48, 58], and the SSN ontology was used to describe sensor data semantically [48, 58, 60]. These ontologies also offered the integration of multiple data types to give a comprehensive approach to KPI calculations. To reuse multiple existing ontologies in one place, these KPI ontologies applied the linked data approach, facilitating interoperability between various domains using semantic web technologies and open protocols. However, the reviewed KPI ontologies only used sensor data rather than dealing with actuator adjustment, whereas this is a required element of an ontology for SOCx of a building.

This review indicated that the main objectives of existing KPI ontologies were to calibrate building models [48] and undertake performance assessments [58, 60]. An important stage in developing a SOCx ontology is to define objectives of evaluating a building's performance. The importance of these objectives can be regarded in terms of assisting building managers and operators in performance failure detection at the building or room level and what can be performed for the better operation of existing buildings. Hence, the principles behind KPIs such as scales (spatial and temporal) and applications (e.g. energy/comfort assessment, building control/optimization, FDD) should be defined in a SOCx ontology development.

## 5.2. Building performance improvement

Reducing building energy use and improving occupants' comfort play a central role in developing a SOCx ontology. Analysis of existing ontologies with this application showed their main objectives as to improve building performance through engaging building users (e.g. facility manager and occupants) [21, 51, 54, 57, 59, 69] or using automatic technologies such as building controls [56, 52] and optimization process [62, 55].

Several ontologies [21, 51] focused on providing contextual information as an important element of an ontology structure for the purpose of engaging building users in building performance improvement. Han et al. [21] proposed a context-aware ontology consisted of a context information aggregator (to collect information), a context reasoner (to extract inferences), and a service reasoner component (to provide a solution) to propose energy saving solutions to building managers based on a simulation-aided process. This ontology was implemented in a single zone of an office building, where five inputs (door, window, fan coil unit, light, and blind) were considered to identify whether the zone had any energy waste using the RDF schema for reasoning, the D2RQ ontology translator [86] to convert raw data into uniform resource identifiers, the OWLIM-RDF database [87] as an ontology repository to load, manage, and read semantic information, and a building performance simulation tool (EnergyPlus) to calculate energy savings from various solutions. This case study showed the capability of the proposed ontology in identifying the energy waste context and energy saving solutions.

Similarly, Kadolsky et al. [51] developed a monitoring system ontology, called InterfaceOnto, comprised of a structure (to include building models), an environment (to contain simulation models), and a control module (to describe context) to recommend required changes in HVAC systems to facility managers using a platform called MonitoringLab. The InterfaceOnto ontology reduced the complexity of the IFC schema-based data, formulated the thermal properties of a building, and defined the selection criteria for improving the energy efficiency of a building. Kadolsky et al. [51] exemplified a filtering method to extract task-specific data from an educational building model at the design phase rather than the operation phase; thus requiring evaluation of the ontology performance in practice.

The ontologies aimed at giving facility managers insights into reducing energy use while maintaining occupants' comfort were Bottaccioli et al.'s [54], the EEPSA [57], and ESIM ontology [69]. Bottaccioli et al. [54] synthesized existing ontologies to make an architecture consisted of a data-sources integration layer (to integrate heterogeneous technologies into web services using IoT gateways), services layer (to develop generic applications and services), and application layer (to develop simulation and visualization) to provide facility managers with potential retrofitting actions. The proposed architecture provided a cloud-hosted software solution, hence facilitating IoT technologies integration. Using existing ontologies for the semantics framework module of the services layer for data relationship inference and sensor information retrieval, the capability of the ontology was demonstrated at the zone level in an educational building. Moreover, the proposed ontology supported calibration of building energy simulation with real-time data. However, it did not support using simulated data outputs for actuators.

The EEPSA ontology [57] integrated BMS data (using the SSN and SOSA ontology) with physical building information (using the BOT ontology) in a space of an educational building in Spain for the manual control of HVAC systems by building operators. This ontology eased building data analysis for building operators through making a semantic relationship between BMS data and building spaces. However, time delays in data registration on the database posed limitations on real-time HVAC system control. The ESIM ontology [69] also combined BIM and BMS data to extract actionable knowledge for facility managers using rule-based semantic reasoning. However, a case study was not provided to evaluate its functionality.

On the occupant front, the OPTIMUS ontology [59] was to reduce building energy use through providing feedback to occupants on their impacts on building performance as a motivation for them to change their behaviors. This ontology provided end-users in a case study with the information on how much energy they used based on building performance assessment, while it simultaneously proposed occupants solutions for energy use reduction and improved comfort. Applying the OPTIMUS ontology in a lab in Athens resulted in the reduction of the energy use and operating cost of the lab compared to the year before using the ontology. This ontology instantiated the inclusion of a wide variety of building data types. The number of data types speaks to the functionality of the ontology since most data types originate from different systems in a building while existing ontologies can be used to define data points and the associated relationships. However, as the number of existing ontologies to combine in a single ontology increases, data point inference becomes more time, effort, and cost intensive.

To improve building performance via automatic building controls, Uribe et al. [52] presented a context-aware architecture for managing sensible heat energy storage systems for which they synthesized existing ontologies to develop an ontology to describe sensors and knowledge-based information in a near zero energy building in Spain. The use case showed that this ontology supported building-related decision making using SPARQL and Semantic Web Rule Language (SWRL) rules. The simplified approach of the ontology allowed for even inexperienced users to obtain information about the building system performance. The BACS ontology [56] is another approach, designed to automatically control window shades in a case study for room-level control based on SPARQL queries and using the proposed EXPRESION ontology for formalizing logical and algebraic expressions, the OSPH ontology for objects' definition, history, state, and unit along with the reuse of the SSN ontology (for BMS data), the BOT ontology (for physical building information data), and the FSM ontology (to control the window shade position).While several existing ontologies were reused in the BACS ontology, some classes (e.g. osph:UnitOfMeasurement) were redefined rather than reusing existing ontologies (e.g. OWL-Time).

For building performance optimization in a swimming pool, an original ontology called SPORTE2 [62] used simulation methods such as artificial neural network and genetic optimization algorithm, real-time sensors and actuators data, and SWRL rules. In the optimization process, an optimal control was chosen (from various predefined optimization scenarios) and simulated to evaluate its feasibility so that changing the associated values in the BMS using the ontology. Schachinger and Kastner [55] also proposed a knowledge-based framework for building energy optimization using machine-readable semantics, for which they developed an ontology to host the required information for building optimization (such as building, environment, and BMS data). They discussed the functionality of their ontology for energy optimization at the zone level, where the information of spaces were extracted efficiently from the ontology using SPARQL for data retrieval; and the expert knowledge defined in the framework were easy accessible as the definitions were explicitly explained in the ontology. SPORTE2 [62] and Schachinger and Kastner's [55] ontologies were the ones among others where the communication between numerical methods and real-time sensors and actuators was an important component of the ontology. Using the ontology-based process for building optimization supports online execution of simulation to improve building performance.

All the reviewed ontologies except for [62, 55] with building performance improvement application were developed based on reusing existing ontologies, promoting their extensibility, reusability, and interoperability. To facilitate ontology and knowledge reuse, several ontologies [21, 51] decomposed ontologies into components. This modularization approach also provides ontology developers with an effective solution for maintaining and extending ontologies. Furthermore, the reviewed ontologies showed that physical building and environmental data were the common data types that commonly used in developing ontologies to provide contextual information and ease real-time data analysis for building users (facility manager, building operator, occupants).

On the other hand, the majority of use cases were at the zone level and a few were at the building level [52, 60], whereas implementation of an ontology at the building level is required to evaluate the ontology performance. Moreover, while the capability of the reviewed ontologies was mostly demonstrated in case studies, the generalizability of the ontologies should be established through implementing them in various contexts. Furthermore, among the reviewed ontologies, the EEPSA ontology [57] was the only case where the time delay in data registration on the database was evaluated while building performance improvement applications (e.g. user interfaces, building controls) are sensitive to latency, hence requiring fast querying [28].

Analysis of ontologies with building performance improvement application showed that the inclusion of contextual information in a SOCx ontology is necessary to equip building users with actionable and meaningful information [21, 51, 52], as they contextualize energy use and occupants' comfort in a building. Beyond providing contextual information, a semantic reasoner [21] is necessary to interpret sensor measurements and infer a solution to building users on how to reduce building energy use. To this end, human-readable medium is also required for conveying the context to building users as they may not have the expertise in the field. Integrating control rules (e.g. using SWRL) with BMS sensors and actuators should be also supported in a SOCx ontology for building performance improvement [56]. However, moving beyond simple rule-based controls to include advanced controls in a SOCx ontology is necessary. Likewise, simulation-based data using advanced simulation techniques (e.g. optimization algorithms and artificial neural network) is a critical part of a SOCx ontology [62, 55].

### 5.3. Fault detection and diagnosis

To perform SOCx of a building, detecting and diagnosing the cause of faults is imperative to ensure that building systems and equipment work properly. Within the reviewed ontologies, several aimed at supporting FDD applications. For example, the CTRLont ontology [35], consisted of context, sense-process-actuate, and control logic parts, facilitated fault detection using simple rule-based controls of HVAC systems. Based on the VDI Guideline [88], control logics were represented graphically using state graphs, which allowed using textual description. The usefulness of the CTRLont ontology was demonstrated in a use case for controlling a virtual (simulated) AHU employing OWL and Unified Modelling Language (UML), where the ontology faciliated detection of a fault emulated in the system. As the CTRLont ontology used OWL, it could be reused in relevant applications. It also provided the explicit specification of control logic using UML state machines, state graphs, and schedules. Moreover, incorporating a modular desing pattern in the CTRLont ontology improved its extensibility and maintainability. However, this ontology is limited to simple control logic rather than advanced building controls such as predictive controls, as there is a gap in how to use open-source ontologies for advanced building controls with respect to the intellectual property (IP) of advanced control models.

A FDD ontology was also proposed by Mallak et al. [68] described the diagnostic rules and relationships between sensors, faults, and HVAC systems using semantic knowledge bases to detect a demand-controlled ventilation (DCV) system's faults in an office building. The consistency in the ontology, which is an important factor in developing ontologies, was checked using HermiT reasoner [89]. Using OWL in this ontology improved its potential for reusability. However, Mallak et al. [68] created rule-based graphs for sensor FDD, rather than a data-driven models as they may be computationally expensive for more complex systems and larger buildings with more spaces.

Another example is Tamani et al.'s [70] ontology on the basis of the FUSE-IT ontology [65] in which the physical and conceptual information about five aspects of a building (assets, spaces, data points, KPIs, and incidents discovery and management) and the relationships between them were defined and semantic technologies were used for anomaly detection in building operation using existing ontologies (SSN, SEAS, SAREF). To demonstrate the functionality of the ontology, it was implemented in two use cases at the zone level: detecting if the number of occupants exceeded the maximum allowable number, and time delays in temperature variations once a mechanical heating system was on. This ontology provided access to data while actuating them for smart building operation in a secure environment for critical buildings such as industrial site, hospital, and public buildings based on the FUSE-IT ontology [65] where four domains of energy, security, facility, and ICT were considered for critical sites.

Analysis of existing FDD ontologies showed that similar to ontologies with building performance improvement applications, contextual information lends itself well to context-based reasoning to detect faults. Communication between sensors and actuators is also a necessary element of FDD ontologies similar to ontologies for building controls. In this regard, automatic inference of the relationships between various data types [76] plays an important role in developing ontologies within the SOCx domain. The reviewed FDD ontologies were limited to simple rules based on the relationship between BMS sensors and actuators. However, inclusion of advanced FDD algorithms in a SOCx ontology is a critical element, requiring future work. In this regard, IP of the advanced models needs to be protected in FDD ontolgies. In addition, a secure access to sensor and actuator data in ontolgoies for smart operation of public buildings is imperative.

## 6 Discussion

This review revealed that the common building data types among existing ontologies were physical building information and BMS data. Physical building information is created once for sensors, building spaces, and devices, and is therefore a static data type [46]. Static data types provide context, allowing for context-aware decision making by SOCx applications. On the other hand, dynamic data types [46] are inevitable data points of a SOCx ontology, as these data types facilitate real-time data analytics to provide actionable knowledge for facility managers. Dynamic data types were present in almost all the reviewed ontologies. BMS data were the most commonly used dynamic data types in ontologies due to the prevalence of BMS-integrated sensors and actuators in existing buildings and their crucial role in SOCx of a building.

In the reviewed ontologies, static data types were typically defined by the IFC schema as this schema is a standard model in defining physical AECO-related data [25]. However, dynamic data types are not well suited for the pervasively used IFC schema [26] because unlike static data, dynamic data vary by time [46]. Among the various ontologies used for dynamic data types, the SSN ontology was frequently reused in existing ontologies as it provides a comprehensive and clear definitions for sensors and actuators of a system, applicable to a wide variety of use cases. However, at present the Brick ontology [27] provides SOCx applications the most comprehensive BMS and contextual information within the building domain.

Brick was developed using a base dataset of more than 17,700 data points in six buildings equipped with BMS from different vendors, and has been demonstrated to capture 98% of these points [29].

To simultaneously cover both static (typically BIM) and dynamic (typically BMS) data types, existing ontologies were reused in 65% of new ones. This is to the benefit of extending existing ontologies when reused by other ontology developers. Furthermore, using existing ontologies promotes a modular ontology design in which elements can be added or removed depending on data domain needs. By not using existing ontologies, developers increase the level of implementation effort. Moreover, it is unlikely that wholly original ontologies to be applicable to other buildings given the ontologies that exist in the field. This is further complicated by the proprietary nature of systems when a new ontology (e.g. [53]) is created from the semantic information of a set of proprietary systems.

Beyond the necessity of developing a coherent data structure, data types required for various applications of a SOCx ontology should also be included to convey a knowledge base to end users in a human readable format. To this effect, on the basis of the reviewed ontologies with the three main applications of KPI calculation, building performance improvement, and FDD, a SOCx ontology should cover KPIs at various spatial and temporal scales for building performance assessment, contextual information for reasoning situations, and control and optimization rules for building performance improvement. However, at present, the reviewed ontologies were limited to simple rules for building controls and FDD, while advanced analytics methods such as artificial neural network and optimization algorithms, should also be deployed in a SOCx ontology.

To share cross-domain data for various SOCx applications, several reviewed ontologies [14, 53, 58, 60] used the linked data approach. Reuse of existing ontologies for combining cross-domain data necessitates implementation of the linked data approach. This approach uses open protocols and semantic web technologies to connect data structured using different ontologies. In contrast, the central data model approach relies on integrating cross-domain data to a common model, which requires converting all data into a single data format [12, 58]. Based on this review, the authors recommend using the linked data approach in developing a SOCx ontology to exchange the data among various stakeholders in the AECO industry as well as to facilitate inclusion of advanced algorithms for building controls and operations.

Irrespective of the data model approach, given the vast amounts of data available in existing buildings, using a query language is required once the cross-domain data is structured based on an ontology. Using a query language has also been incorporated as a method to evaluate the performance of an ontology in returning the data of interest in a reasonable time. This review showed that the common language used to query ontologies for SOCx applications was SPARQL [61, 52, 55, 56, 29, 58, 66], as this query language can effectively support various relationships between structured data [29] to provide access to the appropriate and insightful data for different end users.

It is evident from this review that several gaps remain in the development of ontologies for SOCx applications. A notable gap is the lack of a holistic ontology meeting all SOCx applications. To achieve a holistic SOCx ontology, additional research in data types to support all SOCx applications (KPI calculation, building performance improvement, and FDD) is required. For instance, maintenance, simulation, and performance-based data are required to support SOCx applications while these data types were found in relatively few ontologies (19%, 26%, and 35%, respectively). Before all necessary data types can be included in an ontology, further research in capturing data is also required [53]. For instance, where BIM models do not exist, strategies such as 3D scanning or automated processes to convert 2D floor plans to BIM models [90] can be employed to avoid manually inputting geometric data.

Data relationships are also difficult to define. For example, Borgo et al. [25] noted the difficulty in uncovering the seemingly simple relationships between IFC's "types" and "occurrences" and their ontological representation as classes and instances in OWL. The larger the number of existing ontologies to be included in a proposed ontology, the more complicated this conversion would be. Hence, further research to integrate multiple data types from various existing ontologies in a single ontology is necessary. For instance, the linked data approach [58, 60] is an effective solution in this regard. To minimize manual intervention in the data mapping process, AI-based methods can be greatly beneficial for metadata inference [73, 74] and relation inference [75, 76].

Beyond proposing comprehensive ontologies, evaluations methods such as using ontology scalability and extensibility checking tools [27] for testing proposed ontologies were inadequate in the literature. To check an ontology's completeness, conciseness, and consistency [91], four evaluation approaches were evident in the literature [92]: (1) ontology comparison, (2) domain-specific document comparison, (3) expert review, and (4) case studies. While the first three evaluation approaches were typically not mentioned in papers focused on ontology development, the application case studies were mostly employed to demonstrate the functionality of proposed ontologies. To address this gap in the literature, IT and AECO professionals need to collaborate (e.g. [93]). However, generally, there is a lack of dialogue between these two professionals, whereas interdisciplinary activity of building data ontologies and broadening this research topic would support the nexus of building data ontology and architecture.

The most common shortcoming of the existing ontologies assessed was the lack of a case study. By not providing a case study, the scale of the ontology was difficult to interpret and there was no concrete evidence that the proposed ontology could be applied in a real building context. Likewise, little work has been done to implement proposed ontologies in buildings at scale. In this process, open-source ontologies would be helpful. The implementation of open-source ontologies also means allowing extension and maintenance of existing ontologies. However, there are challenges regarding IP for inclusion of advanced building controls in open-source ontologies to protect creators' rights, especially in commercialization cases. In particular, research should address how to identify operation states automatically advanced control algorithms necessitates without compromising the IP of the developers warrants further investigation [35].

© 2020. Published Article available at
https://www.sciencedirect.com/science/article/abs/pii/S0360132320304741.
This manuscript version is made available under the CC-BY-NC-ND 4.0 license
https://creativecommons.org/licenses/by-nc-nd/4.0/
33

Throughout proposed future research, documentation of a proposed ontology – its vocabularies, data types, specific applications – and the applicability of reused existing ontologies) will be key to the efficiency of SOCx ontology development. Thorough documentation of case studies noting (1) the data types and number of data points; (2) a list of ontologies used, their corresponding applications, and how they were linked to each other; and (3) query languages used would also be of substantial benefit to support future SOCx implementation and allow comparison between different approaches.

# 7    Conclusion

Ontologies developed for general smart building functionality have been the topic of substantial research activity and industry attention in the recent years. However, there has been limited development effort towards a holistic SOCx ontology. This review systematically selected publications relating to data ontology for SOCx published since 2014 and performed a meta-analysis on building data types, ontology approaches, and the overarching applications of ontologies.

This review identified gaps in the literature and the need for further research. Proposed ontologies lacked the ability to meet SOCx ontology application of managing building operation and maintenance data over its operational life, be thoroughly evaluated against their purpose, and be used in future development due to their proprietary nature. Further research should include efforts to add more data types to proposed ontologies, methodologies to accurately capture semantic data for inclusion in ontologies and be managed in an interdisciplinary manner with thorough documentation. Ontology development within the SOCx domain presents an opportunity to make impactful change in reducing the resources required for operation and maintenance while improving comfort throughout the lifecycle of each and every building.

This review paper studied building-scale ontologies, requiring future review of city-scale ontologies. The findings of this paper are expected to be useful for stakeholders in the AECO industry, while a critical review of existing ontologies within the SOCx domain from an IT perspective (e.g. ontology architecture) necessitates future research.

## Acknowledgements

This research was financially supported by the BRAIN Alliance - Big Data Research, Analytics, and Information Network, funded by the Ontario Research Fund Centres of Excellence program and FuseForward, and the Natural Sciences and Engineering Research Council of Canada [CRDPJ 461929 - 13].

[32] P. Arjunan, M. Srivastava, A. Singh and P. Singh, "OpenBAN: An open building analytics middleware for smart buildings," *EAI Endorsed Transactions on Scalable Information Systems,* vol. 2, no. 7, 2015.

[33] M. Arslan, Z. Riaz and S. Munawar, "Building information modeling (BIM) enabled facilities management using Hadoop Architecture," 2017.

[34] L. Kallab, R. Chbeir, P. Bourreau, P. Brassier and M. Mrissa, "HIT2GAP: Towards a better building energy management," *CISBAT 2017 International ConferenceFuture Buildings & Districts – Energy Efficiency from Nano to Urban Scale,* vol. 122, pp. 895-900, 2017.

[35] G. F. Schneider, P. Pauwels and S. Steiger, "Ontology-based modeling of control logic in building automation systems," *IEEE Transactions on Industrial Informatics,* vol. 13, no. 6, pp. 3350-3360, 2017.

[36] B. Yuce and Y. Rezgui, "An ANN-GA semantic rule-based system to reduce the gap between predicted and actual energy consumption in buildings," *IEEE Transactions on Automation Science and Engineering,* vol. 14, no. 3, pp. 1351-1363, 2017.

[37] M. H. Rasmussen, P. Pauwels, C. A. Hviid and J. Karlshøj, "Proposing a central AEC ontology that allows for domain specific extensions," in *2017 Lean and Computing in Construction Congress*, 2017.

[38] S. Rinaldi, A. Flammini, M. Pasetti, L. C. Tagliabue, A. C. Ciribini and S. Zanoni, "Metrological issues in the integration of heterogeneous Iot devices for energy efficiency in cognitive buildings,"
© 2020. Published Article available at
https://www.sciencedirect.com/science/article/abs/pii/S0360132320304741.
This manuscript version is made available under the CC-BY-NC-ND 4.0 license
https://creativecommons.org/licenses/by-nc-nd/4.0/40